\documentclass[11pt,a4paper]{article}
\usepackage[hyperref]{acl2019}
\usepackage{times}
\usepackage{latexsym}
\usepackage{microtype}
\usepackage{booktabs}
\usepackage{multirow}
\usepackage{url}
\usepackage{graphicx}
\usepackage{amsmath}
\usepackage[utf8]{inputenc}
\usepackage{multirow}
\usepackage{xspace}
\newcommand{\eg}{\textit{e.g.}\xspace}
\newcommand{\cf}{\textit{cf.}\xspace}
\newcommand{\ie}{\textit{i.e.}\xspace}
\newcommand{\ia}{\textit{i.a.}\xspace}
\newcommand{\F}{$\text{F}_1$\xspace}
\newcommand{\rt}[1]{\rotatebox{90}{#1}}

\aclfinalcopy
\def\aclpaperid{}

\newcommand{\rot}[1]{\rotatebox{90}{#1}}

\title{Crowdsourcing and Validating\\ Event-focused Emotion Corpora
  for German and English}

\author{Enrica Troiano, Sebastian Pad\'o \and Roman Klinger\\
 Institut f\"ur Maschinelle Sprachverarbeitung \\
  University of Stuttgart, Germany \\
  {\tt \{firstname.lastname\}@ims.uni-stuttgart.de}\\
}

\date{}

\begin{document}
\maketitle
\begin{abstract}
  Sentiment analysis has a range of corpora available across multiple
  languages. For emotion analysis, the situation is more limited,
  which hinders potential research on cross-lingual modeling and the
  development of predictive models for other languages.  In this
  paper, we fill this gap for German by constructing deISEAR, a corpus
  designed in analogy to the well-established English ISEAR emotion
  dataset.  Motivated by Scherer's appraisal theory, we implement a
  crowdsourcing experiment which consists of two steps. In step 1,
  participants create descriptions of emotional events for a given
  emotion. In step 2, five annotators assess the emotion expressed by
  the texts. We show that transferring an emotion classification
  model from the original English ISEAR to the German crowdsourced
  deISEAR via machine translation does not, on average, cause a
  performance drop.
\end{abstract}

\section{Introduction}
Feeling emotions is a central part of the ``human condition''
\citep{russell1945history}. While existing studies on automatic
recognition of emotions in text have achieved promising results
(\newcite{pool2016distant,mohammad2011once}, \ia), we see two main
shortcomings. First, there is shortage of resources for non-English
languages, with few exceptions, like Chinese
\cite{li2017cheavd,Odbal2014,yuan2002acoustic}. This hampers the
data-driven modeling of emotion recognition that has unfolded, \eg,
for the related task of sentiment analysis. Second, emotions can be
expressed in language with a wide variety of linguistic devices, from
direct mentions (\eg, ``\emph{I'm angry}'') to evocative images
(\eg,``\emph{He was petrified}'') or prosody. Computational emotion
recognition on English has mostly focused on explicit emotion
expressions. Often, however, emotions are merely inferable from world
knowledge and experience.  For instance, "\emph{I finally found love}"
presumably depicts a joyful circumstance, while fear probably ensued
when "\emph{She heard a sinister sound}". Attention to such
\textit{event-related emotions} is arguably important for
wide-coverage emotion recognition and has motivated shared tasks
\cite{Klinger2018x}, structured resources \cite{Balahur2011} and
dedicated studies such as the ``International Survey on Emotion
Antecedents and Reactions''
\citep[ISEAR,][]{scherer1994evidence}. ISEAR, as one outcome, provides
a corpus of English descriptions of emotional events for 7 emotions
(anger, disgust, fear, guilt, joy, shame, sadness). Informants were
asked in a classroom setting to describe emotional situations they
experienced. This focus on private perspectives on events sets ISEAR
apart. Even though from psychology, it is now established in natural
language processing as a
textual source of emotional events.

With this paper, we publish and analyze deISEAR, a German corpus of
emotional event descriptions, and its English companion enISEAR, each
containing 1001 instances. We move beyond the original ISEAR in two
respects. (i), we move from on-site annotation to a two-step
crowdsourcing procedure involving description generation and
intersubjective interpretation; (ii), we analyze cross-lingual
differences including a modelling experiment.  Our corpus, available
at \url{https://www.ims.uni-stuttgart.de/data/emotion}, supports the
development of emotion classification models in German and English
including multilingual aspects.

\section{Previous Work}
\label{sect:prev}
For the related but structurally simpler task of sentiment analysis,
 resources have been created in many languages. For German, this
includes dictionaries \cite[\ia]{ruppenhofer2017evaluating}, corpora
of newspaper comments \cite{schabus2017one} and reviews
\cite{Klinger2014a,Ruppenhofer2014,boland2013creating}. Nevertheless,
the resource situation leaves much to be desired.
The situation is even more difficult for emotion analysis. Emotion
annotation is slower and more subjective \citep{Schuff2017}. Further,
there is less agreement on the set of classes to use, stemming from
alternative psychological theories. These include, \eg, discrete
classes vs.\ multiple continuous dimensions
\citep{Buechel2016}. Resources developed by one strand of research can
be unusable for the other \cite{Bostan2018}.

In German, a few dictionaries have been created for dimensional
approaches. Among them is BAWL–R, a list of words rated with arousal,
valence and imageability features \cite{vo2009berlin,
  Briesemeister2011}, where the nouns of the lexicon have been
assigned to emotion intensities, amongst other values.  Still, German
resources are rare in comparison to English ones. To our knowledge,
corpora with sentence-wise emotion annotations are not available for
this language.

In particular, there is no German corpus with speakers' descriptions
of emotionally intense events similar to the English ISEAR. ISEAR, the
``International Survey on Emotion Antecedents and Reactions''
\citep{scherer1997isear}, was conducted by a group of psychologists
who collected emotion data in the form of self-reports. The aim of the
survey was to probe that emotions are invariant over cultures, and are
characterized by patterns of bodily and behavioral changes (\eg,
change in breathing, felt temperature, speech behaviors). In order to
investigate such view, they administered an anonymous questionnaire to
3000 students all over the world, in which participants were asked to
reconstruct an emotion episode associated to one of seven basic
emotions (anger, disgust, fear, guilt, joy, sadness, shame), and to
recall both their evaluation of the stimulus and their reaction to it.
For the final dataset, all the reports were translated to English, and
accordingly, the responses of, \eg, German speakers who took part in the
survey are not available in their original language.

In this paper, we follow \newcite{scherer1997isear} by re-using their
set of seven basic emotions and recreating part of their questionnaire
both in English and German.  In contrast to ISEAR, we account for the
fact that a description can be related to different emotions by its
writer and its readers. Affective analyses have rendered evidence that
emotional standpoints affect the quality of annotation tasks
\citep{buechel2017readers}. For instance, annotation results vary
depending on whether workers are asked if a text is \textit{associated} with an
emotion and if it \textit{evokes} an emotion, with the first phrasing
downplaying the reader's perspective and inducing higher
inter-annotator agreement \citep{mohammad2013crowdsourcing}.  We take
notice of these findings to design our annotation guidelines.

\begin{table*}
\setlength\tabcolsep{4pt}
\centering
\small
\renewcommand{\arraystretch}{1.0}
\begin{tabular}{l|rc r rrrr r rrrr r rrrr r rrr}
\toprule
\multicolumn{2}{c}{}&\multicolumn{1}{l}{Statistics}&&\multicolumn{4}{c}{Temporal Distance}&&\multicolumn{4}{c}{Intensity}&&\multicolumn{4}{c}{Duration}&&\multicolumn{3}{c}{Gender}\\
\cmidrule(lr){3-3}\cmidrule(lr){5-8} \cmidrule(lr){10-13} \cmidrule(lr){15-18} \cmidrule(l){20-22}
\multicolumn{1}{c}{} & Emotion & \#tok  && D & W & M & Y && NV  &M& I & VI && min & h & $>$h & $\geq$d && M & F & O\\
\cmidrule(l){2-2}\cmidrule(lr){3-3}\cmidrule(){5-5}\cmidrule(l){6-6}\cmidrule(l){7-7}\cmidrule(l){8-8}\cmidrule(l){9-9}\cmidrule(l){10-10}\cmidrule(l){11-11}\cmidrule(l){12-12}\cmidrule(l){13-13}\cmidrule(l){14-14}\cmidrule(l){15-15}\cmidrule(l){16-16}\cmidrule(l){17-17}\cmidrule(l){18-18}\cmidrule(l){19-19}\cmidrule(l){20-20}\cmidrule(l){21-21}\cmidrule(l){22-22}
& Anger & 15.1 && 46 & 25 & 31 & 41&& 3 & 25 & 67 & 48 && 23 & 29 & 39 & 52 && 112 & 31 &--\\
& Disgust  & 13.1 && 38 & 38 & 42 & 25 && 12 & 52 & 48 & 31 && 95 & 37 & 8 & 3 && 110 & 33 & --\\
& Fear & 14.0 && 25 & 32 & 37 & 49 && 4 & 24 & 58 & 57 && 50 & 32 & 31 & 30 && 109 & 34 & --\\
&Guilt & 13.8 && 36 & 27 & 30 & 50 && 8 & 57 & 54 & 24 && 41 & 29 & 43 & 30 && 116 & 27 &--\\
& Joy & 11.6 && 40 & 30 & 29 & 44 && 2 & 18 & 60 & 63 && 14 & 18 & 42 & 69 && 107 & 35 & 1\\
\rot{\rlap{~German}} &
Sadness & 11.5 && 29 & 26 & 42 & 46 && 3 & 31 & 43 & 66 && 16 & 9 & 27 & 91 && 113 & 30 &--\\
& Shame & 13.2 && 25 & 28 & 36 & 54 && 24 & 56 & 41 & 22 && 72 & 28 & 24 & 19 && 116 & 27 &--\\
& \textit{Sum} & 13.2 && 239 & 206 & 247 & 309 && 56 & 263 & 371 & 311 && 311  & 182  & 214 & 294 && 783 & 217 & 1\\
\cmidrule(l){2-2}\cmidrule(lr){3-3}\cmidrule(lr){5-8} \cmidrule(lr){10-13} \cmidrule(lr){15-18} \cmidrule(l){20-22}
& Anger  & 28.3 && 45&29 &25 &44 && 9& 34& 48& 52&& 30 &23 &36 & 54&& 62& 81&--\\
& Disgust & 22.4 && 57& 25& 21&40 && 12& 51&37 &43 && 66&27 &24 &26&& 57&86 &--\\
& Fear  & 27.0 && 19 &29 &36 &59 && 2&30 &57 &54 && 52&29 &35 &27 &&66 &77 &--\\
& Guilt & 25.5 && 33 &24 &27 &59 && 25 &52 & 43&23 && 26&39 &28 &50&& 59&84 &--\\
\rot{\rlap{~English}} &
Joy  & 23.6 && 32&24 &31 &56 && 2& 27 & 48 &66 && 14 &13 & 43& 73&& 60&83 &--\\
& Sadness & 21.6 && 40 &24 &31 &48 && 10 & 45 & 38 &50 &&  17&21 &23 &82&& 62&81 &--\\
& Shame & 24.8 && 21&22 &19 &81 && 16&51 &42 &34 && 29&25 &39 &50 && 57 &86 &--\\
& \textit{Sum}  & 24.7 && 247& 177& 190&387 && 76&290 &313 &322 && 234 & 177 & 228 & 362&& 423&578 &--\\
\bottomrule
\end{tabular}
\caption{Statistics for prompting emotions across the average number of tokens
  (\#tok) and the extra-linguistic labels of the descriptions. Temporal Distance, Intensity and Duration report the number of descriptions for events which took place days
  (D), weeks (W), months (M) or years (Y) ago, which caused an
  emotion of a specific intensity (NV: not very intense, M: moderate,
  I: intense, VI: very intense) and duration (min: a few minutes, one
  hour: h, multiple hours: $>$h, one or multiple days $\geq$d); Gender counts of the annotators are reported in the last column
  (male: M, female: F, other: O).}
\label{tab:result}
\end{table*}

\section{Crowdsourcing-based Corpus Creation}
\label{sect:creation}
We developed a two-phase crowdsourcing experiment: one for generating
descriptions, the other for rating the emotions of the descriptions.
Phase~1 can be understood as sampling from
$P(\text{description} | \text{emotion})$, obtaining likely
descriptions for given emotions. Phase~2 estimates
$P(\text{emotion} | \text{description})$, evaluating the association
between a given description and all emotions.  The participants'
intuitions gathered this way are interpretable as a measure for the
interpersonal validity of the descriptions, and as a point of
comparison for our classification results.

The two crowdourcing phases targeted both German and English. This
enabled us to tease apart the effects of the change of setup and
change of language compared to the original ISEAR collection.

\textbf{Phase 1: Generation.}
We used the Figure-Eight (\url{https://www.figure-eight.com})
crowdsourcing platform. Following the ISEAR questionnaire, we
presented annotators with one of the seven emotions in Scherer
and Wallbott's setup, and asked them to produce a textual description
of an event in which they felt that emotion. The task of description
generation was formulated as one of sentence completion (\eg,
``\emph{Ich f{\"u}hlte Freude, als/weil/...}", ``\emph{I felt joy
  when/because ...}"), after observing that this strategy made the job
easier for laypersons, without inducing any restriction on sentence
structure (for details, see Suppl.\ Mat., Section A).
Further, we asked annotators to specify their gender (male, female,
other), the temporal distance of the event (\ie, whether the event
took place days, weeks, months, or years before the time of text
production), and the intensity and duration of the ensuing emotion
(\ie, whether the experience was not very intense, moderately intense,
intense and very intense, and whether it lasted a few minutes, one
hour, multiple hours, or more than one day).  To obtain an English
equivalent to deISEAR, we crowd-sourced the same set of questions in
English, creating a comparable English corpus (enISEAR). The
generation task was published in two slices (Nov/Dec 2018 and Jan
2019). It was crucial for data quality to restrict the countries of
origin (for German, DE/A; for English, UK/IR) -- this prevented a
substantial number of non-native participants who are proficient users
of machine translation services from submitting answers. For each
generated description, we paid 15 cents (see Suppl. Material, Section
A for details).

\textbf{Phase 2: Emotion Labeling.}
To verify to what extent the collected descriptions convey the
emotions for which they were produced, we presented a new set of
annotators with ten randomly sampled descriptions, omitting the
emotion word (\eg, ``\emph{I felt \ldots\ when/because \ldots}''),
together with the list of seven emotions. The task was to choose the
emotion the original author most likely felt during the described
event.  Each description was judged by 5 annotators. We paid 15 cents
per task.

\section{Corpus Analysis}
\textbf{Descriptive analysis.} 
We include all descriptions from Phase~1 in the final resource and the
upcoming discussion, regardless of the inter-annotator agreement from
Phase~2.  Both deISEAR and enISEAR comprise 1001 event-centered
descriptions: deISEAR includes 1084 sentences and 2613 distinct
tokens, with a 0.19 type-token ratio; enISEAR contains 1366 sentences
and a vocabulary of 3066 terms, with a type-token ratio of 0.12.
Table~\ref{tab:result} summarizes the Phase 1 annotation. For each
prompting label\footnote{Transl. de$\rightarrow$en: Angst-Fear,
  Ekel-Disgust, Freude-Joy, Scham-Shame, Schuld-Guilt,
  Traurigkeit-Sadness, Wut-Anger}, we report average description
length, annotators' gender, duration, intensity and temporal distance
of the emotional events.

The main difference between the two languages is description length:
English instances are almost twice as long (24.7 tokens) as German
ones (13.2 tokens). These differences may be related to the
differences in gender distribution between languages.

Most patterns are similar across German and English. In both corpora,
Anger and Sadness receive the longest and shortest descriptions,
respectively. Enraging facts are usually depicted through the specific
aspects that irritated their experiencers, like ``\emph{when a
  superior at work decided to make a huge issue out of something very
  petty just to [...] prove they have power over me}". In contrast,
sad events are reported with fewer details, possibly because they are
often conventionally associated with pain and require little
elaboration, such as ``\emph{my grandmother had passed away}".
Also the perceptual assessments of emotion episodes, as given by the extra-linguistic labels, are comparable between languages. The majority of descriptions are
located at the high end of the scale both for intensity and temporal
distance, \ie, they point to ``milestone'' events that are both remote
and emotionally striking.

\begin{table}
\setlength\tabcolsep{2pt}
\centering
\small
\renewcommand{\arraystretch}{1.0}
\begin{tabular}{l rrrrr l rrrrr}
\toprule
\multicolumn{1}{c}{}&\multicolumn{5}{c}{German} & 
\multicolumn{1}{c}{} & 
\multicolumn{5}{c}{English}\\
\cmidrule(l){2-6}\cmidrule(l){8-12}
Emotion & $\geq$1 & $\geq$2 & $\geq$3&  $\geq$4 & $=$5&
\multicolumn{1}{c}{} & 
 $\geq$1 & $\geq$2 & $\geq$3&  $\geq$4 & $=$5\\
\cmidrule(r){1-1}\cmidrule(l){2-6}\cmidrule(l){8-12}
Anger & 135 & 125 & 107 & 81 & 52&& 137& 129& 112& 89& 59\\
Disgust &139 & 134 & 130  & 124  & 91 && 118& 101& 84& 76&53\\
Fear &134 & 124 & 108 & 99 & 78&& 136&131&124&116& 86\\
Guilt &137 & 126 & 102 & 67 & 31 && 137& 130&124&89& 44\\
Joy &142 & 142 & 142  & 140 & 136 && 143&143& 143&143&137\\
Sadness &132 & 123 & 113 & 97 & 76& &140&133&131&116& 97\\
Shame &128 & 109 & 86 & 66 & 41&& 116&92&64&41&23\\
\cmidrule(r){1-1}\cmidrule(l){2-6}\cmidrule(l){8-12}
\textit{Sum} & 947 & 883 & 788 & 674 & 505 && 927 & 859 & 782& 670 & 499\\ 
\bottomrule
\end{tabular}
\caption{Number of descriptions whose prompting label (column Emotion)
  agrees with the emotion labeled by all Phase-2 annotators ($=$5), by
  at least four ($\geq$4),  at least three ($\geq$3), at least two
  ($\geq$2), at least one ($\geq$1).}
\label{tb:agreement}
\end{table}

\textbf{Agreement on emotions.} 
We next analyze to what extent the emotions labelled in Phase 2 agree
with the prompting emotion presented in Phase 1.  Table
\ref{tb:agreement} reports for how many descriptions (out of 143) the
prompting emotion was selected one, two, three, four, or five (out of
five) times in Phase 2. Agreement is similar between deISEAR
and enISEAR.  This indicates that the German items, although short,
are sufficiently informative.  In both languages, the agreement drops
across the columns, yet half of the descriptions show perfect
intersubjective validity (=5): 505 for German, 499 for English. We
interpret this as a sign of quality.

Again, we find differences among emotions. Agreement is nearly
perfect for Joy and rather low for Shame. These patterns can arise
due to different processes. Certain emotions are easier to recognize
from language (\eg, ``\emph{when I saw someone else got stabbed near
  me}'': Fear) than others (\eg ``\emph{when my daughter was rude to
  my wife}'': elicited for Shame, arguably also associated with Anger
or Sadness). Patterns may also indicate closer conceptual similarity
among specific emotions \cite[\cf]{russell1977evidence}.

\begin{figure}
  \centering
  \includegraphics[scale=0.47,page=1]{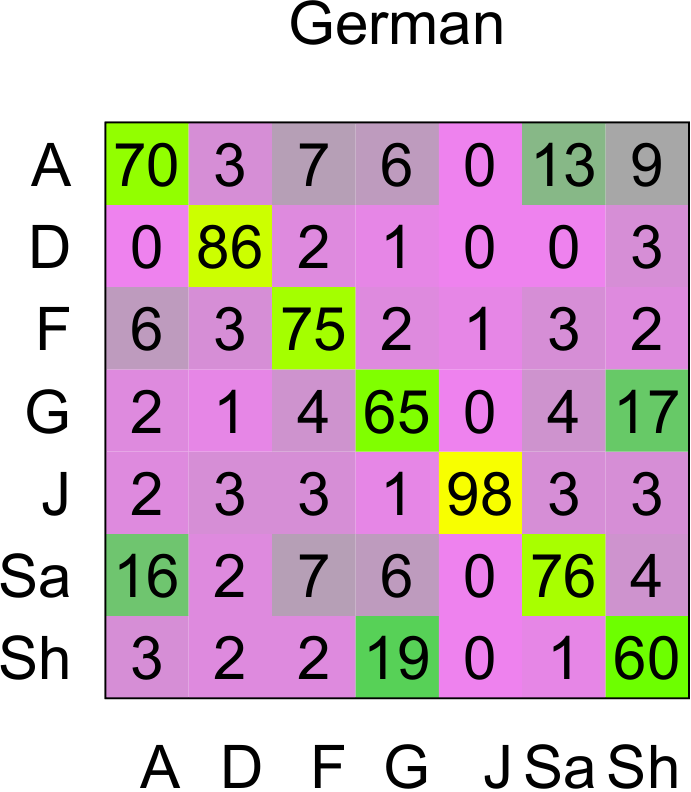} 
  \includegraphics[scale=0.47,page=2]{heatmap} 
  \caption{Confusion matrices for emotions. Columns: prompting
    emotions; rows: labeled emotions.}
  \label{fig:confusionmatrix}
\vspace{-.1cm}
\end{figure}

To follow up on this observation, Figure \ref{fig:confusionmatrix}
shows two confusion matrices for German and English which plot the
frequency with which annotators selected emotion labels (Phase~2,
rows) for prompting emotions (Phase~1, columns).
The results in the diagonals correspond to the =5 columns in
Table~\ref{tb:agreement}, mirroring the overall high level of validity
of the descriptions, and spanning the range between Joy (very high
agreement) and Shame (low agreement). The off-diagonal cells indicate
disagreements. In both languages, annotators perceive Shame
descriptions as expressing Guilt, and vice versa (35\% and 15\% for
English, 17\% and 19\% for German).  In fact, Shame and Guilt
\textquotedblleft occur when events are attributed to internal
causes\textquotedblright\space\citep{tracy2006appraisal}, and thus
they may appear overlapping.

We also see an interesting cross-lingual divergence. In deISEAR,
Sadness is comparably often confused with Anger (13\% of items), while
in enISEAR it is Disgust that is regularly interpreted as Anger (25\%
of items).  This might results from differences in the connotations of
the prompting emotion words in the two languages. For Disgust
(``\emph{Ekel}''), German descriptions concentrate on physical
repulsion, while the English descriptions also include metaphorical
disgust which is more easily confounded with other emotions such as
Anger.

\textbf{Post-hoc Event type analysis.} After the preceding analyses,
we returned to the Phase 1 descriptions and performed a post-hoc
annotation ourselves on a sample of 385 English and 385 German
descriptions (balanced across emotions). We tagged them with
dimensions motivated by \newcite{smith1985patterns}: whether the event
was reoccurring (\textit{general}), whether the event was in the
\textit{future} or in the \textit{past}; whether it was a
\textit{prospective} emotion or actually felt; whether it had a
\textit{social} characteristic (involving other people or animals);
whether the event had \textit{self consequences} or
\textit{consequences for others}; and whether the author presumably
had \textit{situational control} or
\textit{responsibility}\footnote{One may be responsible, but not in
  control of the situation (\eg, ``\emph{when I forgot to set an
    alarm}").}.

\begin{table}[t]
\setlength\tabcolsep{5pt}
\centering
\small
\renewcommand{\arraystretch}{1.0}
\begin{tabular}{ll rrrrrrr}
\toprule
\cmidrule(l){3-9}
&\textit{Dimension} & \rt{Anger} & \rt{Disgust} & \rt{Fear}&  \rt{Guilt} & \rt{Joy}& \rt{Sadness}& \rt{Shame} \\
\cmidrule(l){1-2}\cmidrule(l){3-9}
\multirow{9}{*}{\rt{German}}
&General event  & 4 & 2 & 1 & 0 & 0 & 1 & 0 \\
&Future event   & 0 & 0 & 1 & 0 & 0 & 0 & 0 \\
&Past event     & 51& 53& 53& 55& 55& 54& 55\\
&Prospective    & 1 & 0 &4 & 0 & 1 & 1 & 0 \\
&Social         &30 & 28 & 24 &29 & 24 & 40 &25 \\
&Self conseq.   &37 & 34 & 37 &26 & 44 & 21 &37 \\
&Conseq. oth.    &21 & 9 & 19 &34 & 16 & 34 &14 \\
&Situat. control     & 2 & 5 &4 &24 & 9 & 3 &19 \\
&Responsible    & 20 & 31 & 17 &51 & 26 & 23 &40 \\
\cmidrule{1-2}\cmidrule(l){3-9}
\multirow{9}{*}{\rt{English}}
&General event  & 2 & 2 &2 & 2 & 0 & 3 & 0 \\
&Future event   & 0 & 0 &0 & 0 & 0 & 0 & 0 \\
&Past event     & 53&53 &53&53 &55 &52 & 55 \\
&Prospective    & 0 & 0 & 14 & 0 & 1 & 0 & 0 \\
&Social         & 50 & 37 & 30 &41 & 39 & 49 &41 \\
&Self conseq.   & 29 & 26 & 42 &20 & 35 & 16 &32 \\
&Conseq. oth.   & 29 & 23 & 19 &34 & 24 & 43 &29 \\
&Situat. control     & 3 & 7 &8 &31 & 15 & 2 &24 \\
&Responsible      &13 & 29 & 34 &53 & 34 & 16 &43 \\
\bottomrule
\end{tabular}
\caption{Event type analysis: Cells are counts of post-annotation out
  of 55 descriptions for each emotion.}
\label{tab:OCCraw}
\end{table}

Table~\ref{tab:OCCraw} shows the results.  In both English and German,
only a few units depict general and future events, in line with the
annotation guidelines. Fear more often targets the future than other
emotions. Most event descriptions involve other participants,
especially in English.
In general, events seem to affect authors themselves more than other
people, particularly in the case of Joy and Fear. Exceptions are Guilt
and Sadness, for which there is a predominance of events whose effects
bear down on others.
Regarding the aspect of situational control, Shame and Guilt dominate.
Guilt is particularly more frequent in descriptions in which the
author is presumably responsible. These observations echo the findings
by \newcite{tracy2006appraisal}.

\textbf{Modeling.} As a final analysis, we tested the compatibility of
our created data with the original ISEAR corpus for emotion
classification. We trained a maximum entropy classifier with L2
regularization with boolean unigram features on the original ISEAR
corpus (7665 instances) and evaluated it on all instances collected in
Phase 1 \citep[with liblinear,][]{fan2008liblinear}. We chose MaxEnt
as a method as it constitutes are comparably strong baseline which is,
in contrast to most neural classifiers, more easy to reproduce due to
the convex optimization function and fewer hyper-parameters.  We
applied it to enISEAR and to a version of deISEAR translated with
Google Translate\footnote{\url{http://translate.google.com}, applied
  on February 25, 2019}, an effective baseline strategy for
cross-lingual modeling \citep{Barnes2016}. In accord with the Phase~2
experiment, the emotion words present in the sentences were obscured.
Table~\ref{tab:modeling} shows a decent performance of the ISEAR model
on our novel corpora, with similar scores and performance differences
between emotion classes to previous studies
\citep{Bostan2018}.

Modeling performance and inter-annotator disagreement are correlated:
emotions that are difficult to annotate are also difficult to predict
(Spearman's $\rho$ between \F and the diagonal in Figure
\ref{fig:confusionmatrix} is 0.85 for German, $p=$ .01, and 0.75 for
English, $p=$ .05). It is notable that results for German are on a
level with English despite the translation step and the shorter length
of the German descriptions. That goes against our expectations, as
previous studies showed that translation is only sentiment-preserving
to some degree \cite{Salameh2015,Lohar2018}.  We take this outcome as
evidence for the cross-lingual comparability of deISEAR and enISEAR,
and our general method.

\begin{table}[tb]
    \centering
    \setlength\tabcolsep{3pt}
    \small
    \renewcommand{\arraystretch}{1.0}
    \begin{tabular}{lccccccccc}
    \toprule
        Dataset & $\mu$\F & An & Di & Fe & Gu & Jo & Sa & Sh \\
        \cmidrule(r){1-1}\cmidrule(lr){2-2}\cmidrule(l){3-9}
        deISEAR & 47 & 29 & 49 & 48 & 42 & 68 & 53 & 39 \\
        enISEAR &  47 & 27 & 45 & 57 & 41 & 67 & 58 & 32 \\
    \bottomrule
    \end{tabular}
    \caption{Performance of ISEAR-trained classifier on our
      crowdsourced corpora, per emotion and micro-average \F ($\mu$\F).}
    \label{tab:modeling}
\end{table}

\section{Conclusion}
\label{sect:concl}

We presented (a) deISEAR, a corpus of 1001 event descriptions in
German, annotated with seven emotion classes; and (b) enISEAR, a
companion English resource build analogously, to disentangle effects
of annotation setup and English when comparing to the original ISEAR
resource. Our two-phase annotation setup shows that perceived emotions
can be different from expressed emotions in such event-focused corpus,
which also affects classification performance.

Emotions vary substantially in their properties, both linguistic and
extra-linguistic, which affects both annotation and modeling, while
there is high consistency across the language pair
English--German. Our modeling experiment shows that the
straightforward application of machine translation for model transfer
to another language does not lead to a drop in prediction performance.

\section*{Acknowledgments}
This work was supported by Leibniz
WissenschaftsCampus Tübingen ``Cognitive Interfaces'' and Deutsche
Forschungsgemeinschaft (project SEAT, KL 2869/1-1). We thank Kai
Sassenberg for inspiration and fruitful discussions.

\clearpage

\appendix

\onecolumn

\section{Corpus Generation and Labelling}
For experimental reproducibility, we detail here our crowdsourcing
approach. Figure \ref{fig:instructions} illustrates the instructions
presented to the annotators for sentence generation (Phase 1), Figure
\ref{fig:task1} shows a preview of the task itself. The labelling task
of Phase 2 is presented in Figure \ref{fig:task2}.

To built deISEAR, we targeted Figure-Eight contributors from Germany
and Austria, while the English experiment was restricted to United
Kingdom and Ireland. As a quality check, we required all workers to be
level-3 contributors, i.e., the most experienced ones, who reached the
highest accuracy in previous Figure-Eight jobs. It should be noted
that these laypeople received only minimal and
distant training, while participants of ISEAR were directly instructed
by the experimenters. We aimed at adapting their questionnaire to a
crowdsourcing framework, by formulating the task of sentence
generation as one of sentence completion (e.g. ``\emph{Ich f{\"u}hlte
  Freude, als/weil/...}'', ``\emph{I felt Joy when/because
  ...}''). Preliminary experiments showed that people provided more
coherent and grammatically correct sentences than when they were
presented with a faithful translation of the original survey.

Phase 1 involved 121 English jobs and 116 German jobs after filtering unacceptable answers (e.g. nonsensical items), totalling 2002 tasks
(hits). The two languages required a diverse amount of jobs because
ungrammatical and nonsensical descriptions were (manually)
discarded. In the second Phase, 34 jobs were launched for English and 23 for
German. This way we collected 5005 annotations for each language
(i.e. 5 annotations per description). Overall, data collection and
annotation was finalized in three months. The total cost was 300\$ for Phase 1, and 150\$ for Phase 2.

\begin{center}
\includegraphics[width=\linewidth]{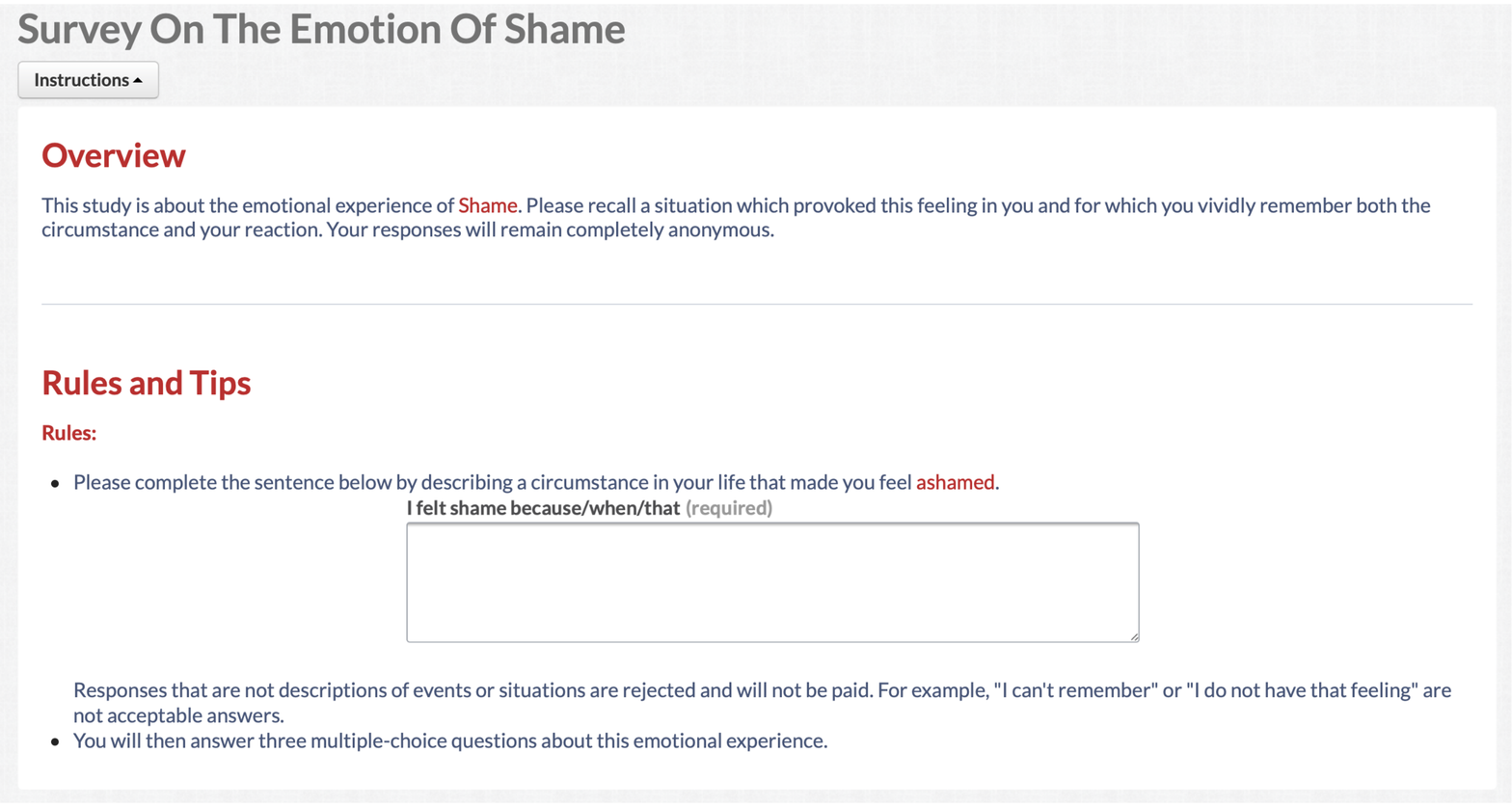}
\captionof{figure}{Instructions for the Generation Task}
\label{fig:instructions}
\end{center}

\begin{center}
  \includegraphics[width=\linewidth]{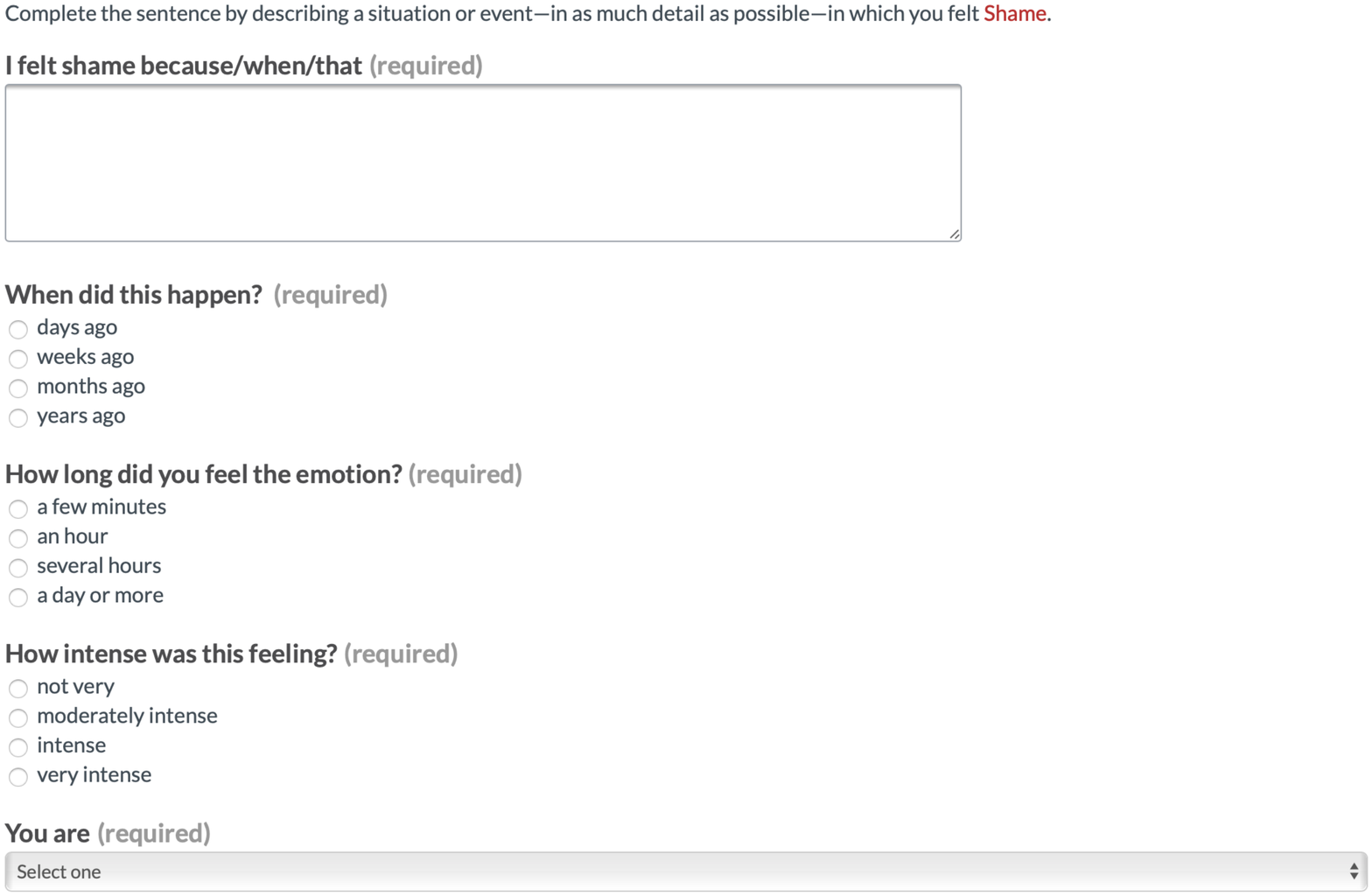}
  \captionof{figure}{Preview of the Generation Task}
  \label{fig:task1}
\end{center}

\begin{center}
  \centering
  \includegraphics[width=\linewidth]{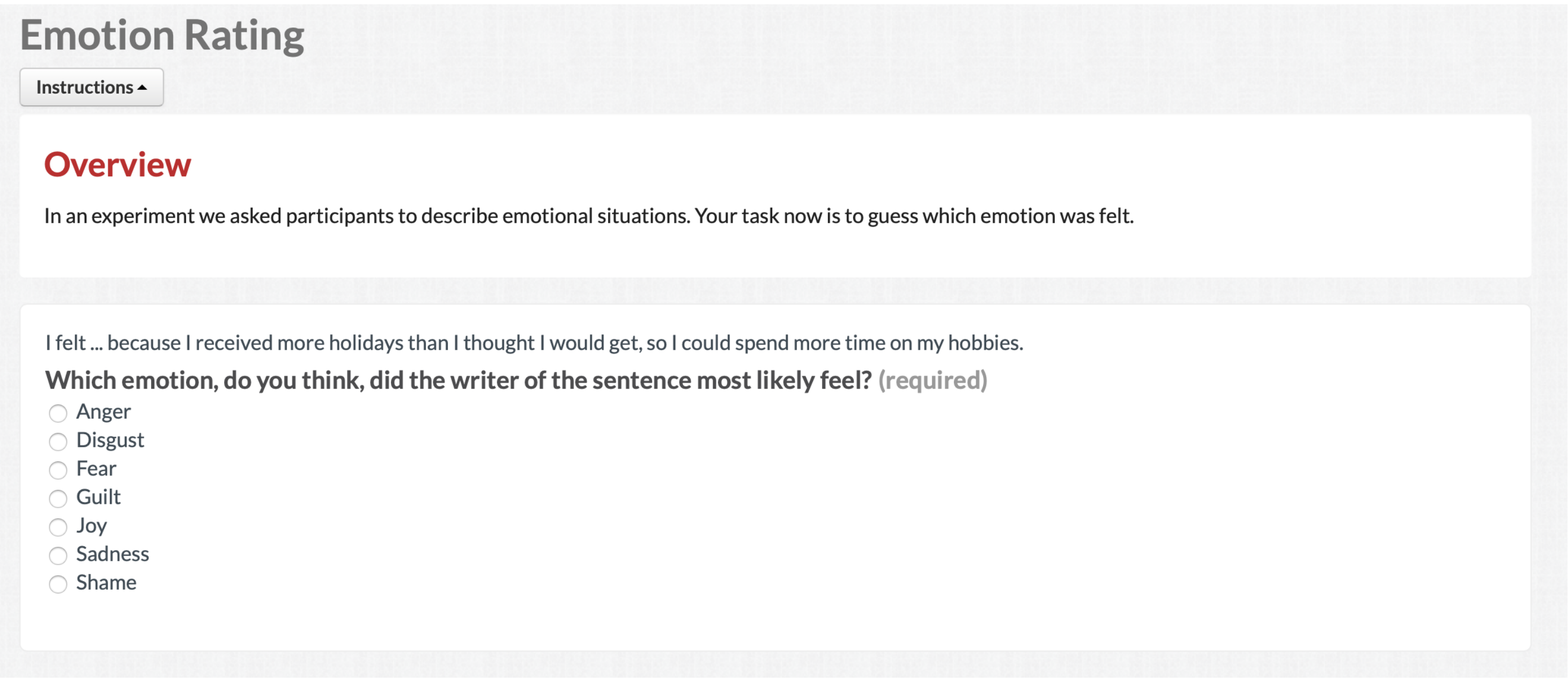}
  \captionof{figure}{Preview of the Emotion Validation Task}
  \label{fig:task2}
\end{center}

\clearpage

\section{Descriptive Analysis}
Table~\ref{tab:description1} and Table~\ref{tab:description2} present
a compact description of the corpora, normalizing the counts by column
and by row blocks, as reported in Section 4 in the main paper.

Table~\ref{tab:description1} highlights differences in the
distribution of emotions across different temporal distances,
intensities, durations, and annotators' gender. We see for instance that
Shame is outstanding in English for long-distant events, while Anger
and Disgust (depending on language) are more dominant in events that
happened a few days prior to description production. For intensities, the distribution across emotions
is most unbalanced for the label ``Not Very''; for duration, Disgust
is the prevailing emotion among those which lasted only a few minutes,
while it is the less frequent among those which persisted for one or multiple days. The exact opposite holds for Joy and Sadness, which
appear to be more durable states.

Table~\ref{tab:description2} highlights differences in the
distribution of extra-linguistic labels across different emotions.  A
few commonalities emerge between the two languages. The majority of
descriptions are referred to remote emotion episodes. Moreover,
Anger-, Fear-, Joy- and Sadness-related descriptions are mostly about
events which caused very intense affective states.  For duration, most
occurrences of Anger and Sadness lasted longer than one day both in
German and English, while Fear episodes are more short-termed,
similar to Disgust.

\begin{center}
\setlength\tabcolsep{4pt}
\small
\renewcommand{\arraystretch}{.9}
\begin{tabular}{l|r r rrrr r rrrr r rrrr r rrr}
\toprule
\multicolumn{2}{c}{}&&\multicolumn{4}{c}{Temporal Distance}&&\multicolumn{4}{c}{Intensity}&&\multicolumn{4}{c}{Duration}&&\multicolumn{3}{c}{Gender}\\
\cmidrule(lr){4-7} \cmidrule(lr){9-12} \cmidrule(lr){14-17} \cmidrule(l){19-21}
\multicolumn{1}{c}{} & Emotion && D & W & M & Y && NV  &M& I & VI && min & h & $>$h & $\geq$d && M & F & O\\
\cmidrule(l){2-2}\cmidrule(l){4-4}\cmidrule(l){5-5}\cmidrule(l){6-6}\cmidrule(l){7-7}\cmidrule(l){9-9}\cmidrule(l){10-10}\cmidrule(l){11-11}\cmidrule(l){12-12}\cmidrule(l){14-14}\cmidrule(l){15-15}\cmidrule(l){16-16}\cmidrule(l){17-17}\cmidrule(l){19-19}\cmidrule(l){20-20}\cmidrule(l){21-21}
& Anger   &&  .19 &  .12 &  .13 &  .13 &&  .05 &  .10 &  .18 &  .15 &&  .07 &  .16 &  .18 &  .18 &&  .14 &  .14 &  0 \\
& Disgust &&  .16 &  .18 &  .17 &  .08 &&  .21 &  .20 &  .13 &  .10 &&  .31 &  .20 &   .04 &   .01 &&  .14 &  .15 &  0 \\
& Fear   &&  .10 &  .16 &  .15 &  .16 &&  .07 &  .09 &  .16 &  .18 &&  .16 &  .18 &   .14 &   .10 &&  .14 &  .16 &  0\\
& Guilt   &&  .15 &  .13 &  .12 &  .16 &&  .14 &  .22 &  .15 &  .08 &&  .13 &  .16 &   .20 &  .10 &&  .15 &  .12 &  0 \\
& Joy    &&  .17 &  .15 &  .12 &  .14 &&  .04 &  .07 &  .16 &  .20 &&  .05 &  .10 &   .20 &   .23 &&  .14 &  .16 &  1\\
\rot{\rlap{~German}} & Sadness &&  .12 &  .13 &  .17 &  .15 &&  .05 &  .12 &  .12 &  .21 &&  .05 &  .05 &   .13 &   .31 &&  .14 &  .14 &  0 \\
& Shame   &&  .10 &  .14 &  .15 &  .17 &&  .43 &  .21 &  .11 &  .07 &&  .23 &  .15 &   .11 &  .06 &&  .15 &  .12 &  0\\

\cmidrule(lr){4-7} \cmidrule(lr){9-12} \cmidrule(lr){14-17} \cmidrule(l){19-21}
& Anger   &&  .18 &  .16 &  .13 &  .11 &&  .12 &  .12 &  .15 &  .16 &&  .13 &  .13 &   .16 &  .15 &&  .15 &  .14 & 0\\
& Disgust &&  .23 &  .14 &  .11 &  .10 &&  .16 &  .18 &  .12 &  .13 &&  .28 &  .15 &  .11 &  .07 &&  .13 &  .15 & 0\\
& Fear   && .08 &  .16 &  .19 &  .15 && .03 &  .10 &  .18 &  .17 &&  .22 &  .16 &  .15 &  .07 && .16 &  .13 & 0\\
& Guilt   &&  .13 &  .14 &  .14 &  .15 &&  .33 &  .18 &  .14 &  .07 &&  .11 &  .22 &  .12 &  .14 &&  .14 &  .15 & 0\\
\rot{\rlap{~English}} & Joy    &&  .13 &  .14 &  .16 &  .14 &&  .03 &  .09 &  .15 &  .20 &&  .06 &  .07 &   .19 &  .20 &&  .14 &  .14 & 0\\
& Sadness &&  .16 &  .14 &  .16 &  .12 &&  .13 &  .16 &  .12 &  .16 &&  .07 &  .12 &   .10 &   .23 &&  .15 &  .14 & 0\\
& Shame   &&  .09 &  .12 &  .10 &  .21 &&  .21 &  .18 &  .13 &  .11 &&  .12 &  .14 &  .17 & .14 &&  .13 &  .15 & 0\\

\bottomrule
\end{tabular}
\captionof{table}{Statistics normalized by column. The unnormalized
  counts are shown in the paper in Table 1.}
\label{tab:description1}
\end{center}

\begin{center}
\setlength\tabcolsep{4pt}
\small
\renewcommand{\arraystretch}{.9}
\begin{tabular}{l|r r rrrr r rrrr r rrrr r rrr}
\toprule
\multicolumn{2}{c}{}&&\multicolumn{4}{c}{Temporal Distance}&&\multicolumn{4}{c}{Intensity}&&\multicolumn{4}{c}{Duration}&&\multicolumn{3}{c}{Gender}\\
\cmidrule(lr){4-7} \cmidrule(lr){9-12} \cmidrule(lr){14-17} \cmidrule(l){19-21}
\multicolumn{1}{c}{} & Emotion && D & W & M & Y && NV  &M& I & VI && min & h & $>$h & $\geq$d && M & F & O\\
\cmidrule(l){2-2}\cmidrule(l){4-4}\cmidrule(l){5-5}\cmidrule(l){6-6}\cmidrule(l){7-7}\cmidrule(l){9-9}\cmidrule(l){10-10}\cmidrule(l){11-11}\cmidrule(l){12-12}\cmidrule(l){14-14}\cmidrule(l){15-15}\cmidrule(l){16-16}\cmidrule(l){17-17}\cmidrule(l){19-19}\cmidrule(l){20-20}\cmidrule(l){21-21}
& Anger   &&  .32 &  .17 &  .22 &  .29 &&  .02 &  .17 &  .47 &  .34 &&  .16 &  .20 &   .27 &   .36 &&  .78 &  .22 &  0\\
& Disgust &&  .27 &  .27 &  .29 &  .17 &&  .08 &  .36 &  .34 &  .22 &&  .66 &  .26 &   .06 &   .02 &&  .77 &  .23 &  0\\
& Fear    &&  .17 &  .22 &  .26 &  .34 &&  .03 &  .17 &  .41 &  .40 &&  .35 &  .22 &   .22 &   .21 &&  .76 &  .24 & 0\\
& Guilt   &&  .25 &  .19 &  .21 &  .35 &&  .06 &  .40 &  .38 &  .17 &&  .29 &  .20 &   .30 &   .21 &&  .81 &  .19 & 0\\
& Joy     &&  .28 &  .21 &  .20 &  .31 &&  .01 &  .13 &  .42 &  .44 &&  .10 &  .13 &   .29 &   .48 &&  .75 &  .24 &  .01\\
\rot{\rlap{~German}} & Sadness &&  .20 &  .18 &  .29 &  .32 &&  .02 &  .22 &  .30 &  .46 &&  .11 &  .06 &   .19 &   .64 &&  .79 &  .21 & 0\\
& Shame   &&  .17 &  .20 &  .25 &  .38 &&  .17 &  .39 &  .29 &  .15 &&  .50 &  .20 &   .17 &   .13 &&  .81 &  .19 & 0\\

\cmidrule(lr){4-7} \cmidrule(lr){9-12} \cmidrule(lr){14-17} \cmidrule(l){19-21}
& Anger   &&  .31 &  .20 &  .17 &  .31 &&  .06 &  .24 &  .34 &  .36 &&  .21 &  .16 &   .25 &   .38 &&  .43 &  .57 & 0\\
& Disgust &&  .40 &  .17 &  .15 &  .28 &&  .08 &  .36 &  .26 &  .30 &&  .46 &  .19 &   .17 &   .18 &&  .40 &  .60 & 0\\
& Fear    &&  .13 &  .20 &  .25 &  .41 &&  .01 &  .21 &  .40 &  .38 &&  .36 &  .20 &   .24 &  .19 &&  .46 &  .54 & 0\\
& Guilt   &&  .23 &  .17 &  .19 &  .41 &&  .17 &  .36 &  .30 &  .16 &&  .18 &  .27 &   .20 &   .35 &&  .41 &  .59 & 0\\
\rot{\rlap{~English}} & Joy     &&  .22 &  .17 &  .22 &  .39 &&  .01 &  .19 &  .34 &  .46 &&  .10 &  .09 &  .30 &  .51 &&  .42 &  .58 & 0\\
& Sadness &&  .28 &  .17 &  .22 &  .34 &&  .07 &  .31 &  .27 &  .35 &&  .12 &  .15 &  .16 &  .57 &&  .43 &  .57 & 0\\
& Shame   &&  .15 &  .15 &  .13 &  .57 &&  .11 &  .36 &  .29 &  .24 &&  .20 &  .17 &   .27 &   .35 &&  .40 &  .60 & 0\\
\bottomrule
\end{tabular}
\captionof{table}{Statistics normalized by partial row. The unnormalized
  counts are shown in the paper in Table 1.}
\label{tab:description2}
\end{center}

\newpage

\section{Event-type Analysis} 
The event-type analysis presented in Section 4 targeted 385 items per
language (55 descriptions per emotion). Table 2 in the paper shows the
counts of instances associated to the psychological labels across the
seven emotions.

For each description, we annotated the following boolean variables:
\begin{itemize}
\item About the event time:
  \begin{itemize}
  \item Does the text describe a \textit{general event}?
  \item Does the text describe a \textit{future event}?
  \item Does the text describe a \textit{past event}?
  \end{itemize}
\item About the realization of the emotion:
  \begin{itemize}
  \item Is it an actual or a \textit{prospective} emotion?
  \end{itemize}
\item About the embedding in a social environment:
  \begin{itemize}
  \item Are other people or animals part of the event description; is
    it a \textit{social} event description?
  \end{itemize}
\item About the consequences of the event:
  \begin{itemize}
  \item Are there \textit{self-consequences}?
  \item Are there \textit{consequences for others}?
  \end{itemize}
\item About the control of the writer:
  \begin{itemize}
  \item Is the author presumably under \textit{situational control}?
  \item Does the author presumably have \textit{self control/responsibility}?
  \end{itemize}
\end{itemize}

While the paper describes the distribution of labels by emotion, here
we expand the discussion to the extra-linguistic information collected
in Phase 1. Table \ref{tab:occlabelraw} distributes the raw counts
across the annotation values. It should be noticed that the random
descriptions used for this analysis were not balanced with respect to
their values of each variable. For
this reason, Table \ref{tab:occ_normbycol} reports relative counts
(i.e. counts of descriptions normalized by the number of instances
within the label Day, Week, Month etc.).

Some regularities can be observed cross all columns of Table
\ref{tab:occ_normbycol}. For instance, events which involved a
purposeful participation of their experiencer are a minority in both
languages (Sit.~control), and approximately 50\% of the
descriptions mention individuals other than the writer (Social). The
latter proportion, however, is higher for English than for German.

Events that are linked to consequences for the self mostly come from
the German sample (Self conseq.). In German, moreover, such type of
events are recalled more frequently than events that had consequences
on others (Conseq. oth.). The opposite is true for English: emotions
of English authors often wrote about events that affected the life of other people or animals. This
holds irrespective of the temporal distance, the intensity, the
duration of the experience and the gender of the
experiencer. Exceptions are English descriptions of facts which only
lasted a few minutes, and which appear to bring consequences for the
self more than for others (Self~conseq. and Conseq.~oth.
in column min).

As for the responsibility of events, this label is consistent
across all columns in the German sample. Instead, in English we
observe some marked differences. Emotions with a low intensity (column
NV) followed an event which was directly triggered by their
experiencer, but very intense emotions are less frequently associated
to responsibility (column VI). Lastly, shorter events (min) imply the
responsibility dimension more than long ones ($\geq$d).

\newpage

\begin{center}
\setlength\tabcolsep{4pt}
\centering
\small
\renewcommand{\arraystretch}{.9}
\begin{tabular}{l|l r rrrr r rrrr r rrrr r rrr}
\toprule
\multicolumn{2}{c}{}&&\multicolumn{4}{c}{Temporal Distance}&&\multicolumn{4}{c}{Intensity}&&\multicolumn{4}{c}{Duration}&&\multicolumn{3}{c}{Gender}\\
\cmidrule(lr){4-7} \cmidrule(lr){9-12} \cmidrule(lr){14-17} \cmidrule(l){19-21}
\multicolumn{1}{c}{} & Dimension && D & W & M & Y && NV  &M& I & VI && min & h & $>$h & $\geq$d && M & F & O\\
\cmidrule(l){2-2}\cmidrule(l){4-4}\cmidrule(l){5-5}\cmidrule(l){6-6}\cmidrule(l){7-7}\cmidrule(l){9-9}\cmidrule(l){10-10}\cmidrule(l){11-11}\cmidrule(l){12-12}\cmidrule(l){14-14}\cmidrule(l){15-15}\cmidrule(l){16-16}\cmidrule(l){17-17}\cmidrule(l){19-19}\cmidrule(l){20-20}\cmidrule(l){21-21}

& General Event && 2 &3 & 1 &2 && 0 &1 & 4 & 3 && 4 & 0 & 1 &   3 && 6 & 2 & 0 \\
& Future Event && 0 & 0 & 1 & 0 && 0 & 0 & 1 & 0 && 0 & 0 & 1 & 0 && 1 & 0 & 0 \\
& Past Event && 98 & 76 & 101 & 101 && 22 & 92 & 141  & 121 && 121 & 66 & 83 & 106 && 287 & 89 & 0\\
& Prospective && 3 & 2 & 2 & 0 && 0 & 3 & 3 &1 && 2 & 2 & 1 & 2 && 5 & 2 & 0 \\
& Social && 55 & 41 & 53 & 51 && 13 & 43 & 80 & 64 && 70 & 32 & 42 & 56 && 152 & 48 & 0 \\
& Self conseq. &&  54 & 45 & 70 &67 &&15 & 52 &94 &75 && 74 & 36 & 59 & 67 &&   176 &      60 &      0 \\
& Conseq. oth.  && 42 & 30 & 34 & 41 && 10 & 35 & 54 & 48 &&        52 & 25 & 28 & 42 && 110 & 37 & 0 \\
\rot{\rlap{~German}}& Sit. ctrl. && 17 & 13 & 18 & 18 &&         2 & 17 & 29 & 18 && 21 & 10 & 14 & 21 && 56 & 10 & 0 \\
& Responsib.  && 53 & 37 & 63 & 55 && 11 & 57 & 76 & 64 && 68 & 40 &  45 & 55 && 160 & 48 & 0 \\
&\textit{Sum} && 226 & 171 & 242 & 234 && 51 & 208 & 341 &           273 && 291 & 145 & 191 & 246 && 666 & 207 & 0 \\

\cmidrule(lr){4-7} \cmidrule(lr){9-12} \cmidrule(lr){14-17} \cmidrule(l){19-21}

&General Event &&   6 & 2 & 2 & 1 && 2 &2 & 5 & 2 && 5 &  2 &  1 &  3 &&  3 & 8 & 0 \\
&Future Event && 0 &  0 &   0 &  0 &&  0 & 0 & 0 & 0 && 0 & 0 & 0 & 0 &&  0 &   0 &  0 \\
& Past Event && 88 & 61 & 73 & 152 && 23 & 104 & 122  & 125 && 76 & 74 & 85 & 139 && 155 & 219 & 0\\
&Prospective &&  3 &  4 &   3 &  5 && 0 &  2 & 8 &  5 &&  5 & 4 &  3 & 3 && 7 & 8 & 0 \\
&Social  && 73 & 51 & 56 & 107 && 14 & 72 & 94 & 107 && 49 & 52 &  66 &  120 && 103 & 184 & 0 \\
&Self conseq. && 46 & 30 & 34 & 90 && 14 & 57 & 71 & 58 && 52 & 32 & 47 & 69 && 89 & 111 & 0 \\
&Conseq. oth. && 51 & 38 &  39 & 73 && 8 & 49 & 69 & 75 && 30 & 47 & 40 & 84 && 73 & 128 & 0 \\
\rot{\rlap{~English}}& Sit. ctrl. && 15 & 17 & 16 & 42 && 12 & 30 & 25 & 23 && 21 & 19 & 17 & 33 &&    40 &  50 & 0 \\
&Responsib. &&  50 & 36 & 47 & 89 && 20 & 71 & 80 & 51 && 57 & 50 & 53 & 62 && 104 & 118 & 0 \\
&\textit{Sum} && 244 & 178 & 197 & 407 && 70 & 283 & 352 & 321 &&   219 & 206 & 227 & 374 && 419 & 607 & 0 \\
\bottomrule
\end{tabular}
\captionof{table}{Event-type analysis: Raw counts of the labels which
  were manually assigned to a subset of enISEAR and deISEAR, across
  the extra-linguistic information collected in Phase 1. See the text
  for the explanation of variables.}
\label{tab:occlabelraw}

\end{center}

\begin{center}
\setlength\tabcolsep{4pt}
\centering
\small
\renewcommand{\arraystretch}{.9}
\begin{tabular}{l|l r rrrr r rrrr r rrrr r rrr}
\toprule
\multicolumn{2}{c}{}&&\multicolumn{4}{c}{Temporal Distance}&&\multicolumn{4}{c}{Intensity}&&\multicolumn{4}{c}{Duration}&&\multicolumn{3}{c}{Gender}\\
\cmidrule(lr){4-7} \cmidrule(lr){9-12} \cmidrule(lr){14-17} \cmidrule(l){19-21}
\multicolumn{1}{c}{} & Dimension && D & W & M & Y && NV  &M& I & VI && min & h & $>$h & $\geq$d && M & F & O\\
\cmidrule(l){2-2}\cmidrule(l){4-4}\cmidrule(l){5-5}\cmidrule(l){6-6}\cmidrule(l){7-7}\cmidrule(l){9-9}\cmidrule(l){10-10}\cmidrule(l){11-11}\cmidrule(l){12-12}\cmidrule(l){14-14}\cmidrule(l){15-15}\cmidrule(l){16-16}\cmidrule(l){17-17}\cmidrule(l){19-19}\cmidrule(l){20-20}\cmidrule(l){21-21}

& General Event &&  .02 &   .04 &.01 &   .02 &&  0 &.01 & .03 &  .02 &&  .03 &  0 &   .01 & .03 &&  .02 &.02 &0 \\
& Future Event&&  0 &   0 &.01 &   0 &&  0 &0 & .01 &  0 &&  0 &  0 &   .01 & 0 &&  0 &0 &0 \\
& Past Event && .98 & .96 & .98 & .98 && 1 & .99 & .97 & .98 && .97 & 1 & .98 & .97 && .98 & .98 & 0 \\
& Prospective &&  .03 &   .03 &.02 &   0 &&  0 &.03 & .02 &  .01 &&  .02 &  .03 &   .01 & .02 &&  .02 &.02 &0 \\
& Social  &&  .55 &   .52 &.51 &   .50 &&  .59 &.46 & .55 &  .52 &&  .56 &  .48 &   .49 & .51 &&  .52 &.53 &0 \\
\rot{\rlap{~German}}& Self conseq. &&  .54 &   .57 &.68 &   .65 &&  .68 &.56 & .64 &  .60 &&  .59 &  .55 &   .69 & .61 &&  .60 &.66 &0 \\
& Conseq. oth.&&  .42 &   .38 &.33 &   .40 &&  .45 &.38 & .37 &  .39 &&  .42 &  .38 &   .33 & .39 &&  .37 &.41 &0 \\
& Sit. ctrl. &&  .17 &   .16 &.17 &   .17 &&  .09 &.18 & .20 &  .15 &&.17 &  .15 &   .16 & .19 &&  .19 &.11 &0 \\
& Responsib.  &&  .53 &   .47 &.61 &   .53 &&  .50 &.61 & .52 &  .52 &&  .54 &  .61 &   .53 & .50 &&  .54 &.53 &0 \\

\cmidrule(lr){4-7} \cmidrule(lr){9-12} \cmidrule(lr){14-17} \cmidrule(l){19-21}

& General Event   &&   .06 &.03 & .03 &.01 &&   .08 & .02 &  .04 &   .02 &&   .06 &   .03 &.01 &  .02 &&  .02 & .04 &0 \\
& Future Event&&   0 &0 & 0 &0 &&   0 & 0 &  0 &   0 &&   0 &   0 &0 &  0 &&  0 & 0 &0 \\
& Past Event && .94 & .97 & .97 & .99 && .92 & .98 & .96 & .98 && .94 & .97 & .99 & .98 && .98 & .96  & 0\\
& Prospective&&   .03 &.06 & .04 &.03 &&   0 & .02 &  .06 &   .04 &&   .06 &   .05 &.03 &  .02 &&  .04 & .04 &0 \\
& Social  &&   .78 &.81 & .75 &.70 &&   .56 & .68 &  .74 &   .84 &&   .60 &   .68 &.77 &  .85 &&  .65 & .81 &0 \\
& Self conseq. &&   .49 &.48 & .45 &.59 &&   .56 & .54 &  .56 &   .46 &&   .64 &   .42 &.55 &  .49 &&  .56 & .49 &0 \\
\rot{\rlap{~English}}& Conseq. oth.&&   .54 &.60 & .52 &.48 &&   .32 & .46 &  .54 &   .59 &&   .37 &   .62 &.47 &  .59 &&  .46 & .56 &0 \\
& Sit. ctrl.&&   .16 &.27 & .21 &.27 &&   .48 & .28 &  .20 &   .18 &&   .26 &   .25 &.20 &  .23 &&  .25 & .22 &0 \\
& Responsib. &&   .53 &.57 & .63 &.58 &&   .80 & .67 &  .63 &   .40 &&   .70 &   .66 &.62 &  .44 &&  .66 & .52 &0 \\

\bottomrule
\end{tabular}
\captionof{table}{Event-type analysis: Counts are normalized by
  instances with the particular value, e.g., the count in the cell ``Time General''--``D'' is
  normalized by the number of all instances with the associated value
  D (temporal distance of days).}
\label{tab:occ_normbycol}
\end{center}

\newpage

\section{Annotator Agreement}
Section 4.1 discussed the agreement reached by different subsets of
annotators \textit{at each generation
  label}. We report relative counts in Table \ref{tab:agreementnorm} and we extend the analysis in Table
\ref{tb:deISEARagreement}, summing over the prompting emotions. This table shows the
interannotator agreement of Phase-2 annotators with respect to the
meta-information given by the participants of Phase 1, i.e.,
all the alternatives for gender, intensity, duration and temporal distance under the column Labels.

These numbers represent the count of descriptions within a corpus -- and
\textit{not within a generation label}, for which the annotation label
is the same as the generation label. One can read the table as
follows: 177 descriptions from deISEAR, which were labeled as VI by
Phase 1 annotators, were then labelled by 5 Phase 2 annotators with
their original prompting emotion; 506 instances provided by female
annotators for enISEAR were labelled by at least 2 Phase 2 annotators
with their original prompting emotion, and so on.

Notably, in the table of Section 4.2, the maximum value that each cell
can reach is 143, i.e., the total number of descriptions prompted by a
specific emotion. Here, the maximum value varies by cell, because each
meta-data label is assigned to a different number of
descriptions\footnote{For an overview of the distribution of meta-data
  labels over the descriptions, refer to Section 4.1.}. Accordingly,
higher counts do not necessarily indicate stronger agreement.

\begin{center}
\centering
\small
\begin{tabular}{ll rrrrr l rrrrr}
\toprule
\multicolumn{1}{c}{}&\multicolumn{5}{c}{German} & 
\multicolumn{1}{c}{} & 
\multicolumn{5}{c}{English}\\
\cmidrule(l){2-6}\cmidrule(l){8-12}
Emotion & $\geq$1 & $\geq$2 & $\geq$3&  $\geq$4 & $=$5&
\multicolumn{1}{c}{} & 
 $\geq$1 & $\geq$2 & $\geq$3&  $\geq$4 & $=$5\\
\cmidrule(r){1-1}\cmidrule(l){2-6}\cmidrule(l){8-12}
Anger   & .94 &  .87 &  .75 &  .57 &  .36 &&  .96 &  .90 &  .78 &  .62 &  .41 \\
Disgust & .97 &  .94 &  .91 &  .87 &  .64  &&  .83 &  .71 &  .59 & .53 &  .37 \\
Fear&  .94 &  .87 &  .76 &  .69 & .55 &&  .95 & .92 &  .87 &  .81 &  .60 \\
Guilt   &  .96 &  .88 &  .71 &  .47 &  .22 &&  .96 &  .91 &  .87 &  .62 &  .31 \\
Joy &  .99 &  .99 &  .99 &  .98 &  .95 &&  1 &  1 &  1 &  1 &  .96 \\
Sadness &  .92 &  .86 &  .79 &  .68 &  .53 &&  .98 &  .93 &  .92 &  .81 &  .68 \\
Shame   &  .90 &  .76 &  .60 &  .46 &  .29 &&  .81 &  .64 &  .45 &  .29 &  .16 \\
\cmidrule(r){1-1}\cmidrule(l){2-6}\cmidrule(l){8-12}
\textit{Sum} & 6.62 & 6.17 & 5.51 & 4.71 & 3.53 && 6.48 & 6.01 & 5.47& 4.69 & 3.49\\ 
\bottomrule
\end{tabular}
\captionof{table}{Relative agreement counts.}
\label{tab:agreementnorm}
\end{center}

\begin{center}
\centering
\small
\begin{tabular}{l|crrrrr c rrrrr}
\toprule
\multicolumn{2}{c}{}&\multicolumn{5}{c}{German}&\multicolumn{1}{c}{}&\multicolumn{5}{c}{English}\\
\cmidrule(){3-7}\cmidrule(){9-13}
\multicolumn{1}{c}{}& \small{Labels}& $\geq$1 & $\geq$2 & $\geq$3&  $\geq$4 & $=$5&&$\geq$1 & $\geq$2 & $\geq$3&  $\geq$4 & $=$5\\
\cmidrule(l){2-2}\cmidrule(l){3-3} \cmidrule(l){4-4}\cmidrule(l){5-5}\cmidrule(l){6-6}\cmidrule(l){7-7}\cmidrule(l){9-9}\cmidrule(l){10-10} \cmidrule(l){11-11}\cmidrule(l){12-12}\cmidrule(l){13-13}
\multirow{4}{*}{\rot{When}} & D & 226 & 157 & 184  & 209 & 226 && 229 & 211 &  189 & 161 & 115\\
& W & 197 &184 &169&143&108&& 168& 152& 137& 112& 79\\
 & M  & 229 &215 & 198 & 174 & 125 && 177& 165& 154& 138& 109\\
& Y & 295 & 275 & 237 & 200 & 161 && 353& 331& 302&  259& 196\\
\hline
\multirow{4}{*}{\rot{Length}}& min  & 291 & 275 & 245  & 213 & 145 && 223&208&185&162&115\\
& h & 173 & 162 & 151 & 127 & 99 && 162&145&130&106&74\\
 & $>$h & 205 &188& 164 & 139 & 103 && 210&197&178&158&118\\
& $\geq$d & 278 &258 & 228 & 195& 158 && 332&309&289&244&192\\
\hline
\multirow{4}{*}{\rot{Intense}}& NV  & 52 & 46 & 38  & 32  & 18 && 74 & 69& 61 & 51 & 31\\
& M & 241 & 224 & 194 & 162 & 113 && 264&240&217&185&128\\
 & I  & 352 & 331 & 301 & 255 & 197 && 288&267&247&213&165\\
& VI  & 302 &282 & 255  & 225 & 177 && 301&283&257&221&172\\
\hline
\multirow{3}{*}{\rot{Gender}}& M  & 738 & 684 & 604  & 510  & 392 && 386&353&316&273&200\\
& F& 208 & 198 & 183  & 163  & 112 && 541&506&466&397&299\\
 & O  & 1 &1 & 1  & 1  & 1 && -- & -- & -- & -- & --\\
\bottomrule
\end{tabular}
\captionof{table}{Full agreement information for both German and English crowd-sourced corpora.}
\label{tb:deISEARagreement}
\end{center}

\newpage

\section{Modeling}
Table~\ref{tb:classification} shows
the results of the maximum entropy classifier across all emotions.

\begin{center}
\centering\small
\begin{tabular}{l rrrrrr l rrrrrr}
\toprule
\multicolumn{1}{c}{}&\multicolumn{5}{c}{deISEAR} & 
\multicolumn{1}{c}{} & 
\multicolumn{5}{r}{enISEAR}\\
\cmidrule(l){2-7}\cmidrule(l){9-14}
Emotion & TP & FP & FN&  P & R& F1 & \multicolumn{1}{c}{} & TP & FP & FN&  P & R &F1\\
\cmidrule(r){1-1}\cmidrule(l){2-7}\cmidrule(l){9-14}
Anger & 29& 30&114&.49&.20 & .29&& 27 & 32 & 116 & .46 & .19 &.27\\
Disgust &65 & 57 & 78  & .53 & .45 &.49&& 67& 85& 76& .44&.47&.45\\
Fear &70 & 77 & 73 & .48 & .49&.48&& 85&69&58&.55& .59&.57\\
Guilt &75 & 140 & 68 & .35 &.52 &.42&& 79& 161&64&.33& .55&.41\\
Joy &106 & 61 & 37  & .63 & .74 &.68&& 94&43& 49&.69&.66&.67\\
Sadness &63 & 31 & 80 &. 67 & .44 &.53&& 70& 29 & 73&.71& .49&.58\\
Shame &66 & 131 & 77 & .34 & .46&.39&& 49&111&94&.31&.34&.32\\
\cmidrule(r){1-1}\cmidrule(l){2-7}\cmidrule(l){9-14}
\textit{Micro} & 474&527&527&.47&.47&.47 && 471&530&530&.47&.47&.47\\ 
\bottomrule
\end{tabular}
\captionof{table}{Classification results for both corpora.}
\label{tb:classification}
\end{center}

\end{document}